\pdfoutput=1
\documentclass[11pt]{article}

\usepackage[]{acl}

\usepackage{makecell}
\usepackage{times}
\usepackage{latexsym}
\usepackage{svg}
\usepackage{graphicx}
\usepackage{listings}
\usepackage{xcolor} % for colors in code
\usepackage{array}
\usepackage{multirow}
\usepackage{natbib}
\usepackage[T1]{fontenc}
\usepackage[utf8]{inputenc}

\usepackage{microtype}

\usepackage{inconsolata}

\usepackage{graphicx}
\usepackage{float}

\title{L3Cube-IndicQuest: A Benchmark Question Answering Dataset for Evaluating Knowledge of LLMs in Indic Context}

\author{
 \textbf{Pritika Rohera\textsuperscript{1,3}},
 \textbf{Chaitrali Ginimav\textsuperscript{1,3}},
 \textbf{Akanksha Salunke\textsuperscript{1,3}},
 \textbf{Gayatri Sawant\textsuperscript{1,3}}, and
 \textbf{Raviraj Joshi\textsuperscript{2,3}}
\\
 Pune Institute of Computer Technology\textsuperscript {1} \\
 Indian Institute of Technology Madras\textsuperscript {2} \\
 L3Cube Labs, Pune\textsuperscript{3}
}

\begin{document}
\maketitle
\begin{abstract}
Large Language Models (LLMs) have made significant progress in incorporating Indic languages within multilingual models. However, it is crucial to quantitatively assess whether these languages perform comparably to globally dominant ones, such as English. Currently, there is a lack of benchmark datasets specifically designed to evaluate the regional knowledge of LLMs in various Indic languages. In this paper, we present the L3Cube-IndicQuest, a gold-standard factual question-answering benchmark dataset designed to evaluate how well multilingual LLMs capture regional knowledge across various Indic languages. The dataset contains 200 question-answer pairs, each for English and 19 Indic languages, covering five domains specific to the Indic region. We aim for this dataset to serve as a benchmark, providing ground truth for evaluating the performance of LLMs in understanding and representing knowledge relevant to the Indian context. The IndicQuest can be used for both reference-based evaluation and LLM-as-a-judge evaluation. The dataset is shared publicly at \url{https://github.com/l3cube-pune/indic-nlp}.
\end{abstract}
\begin{figure}
    \centering
    \includegraphics[width=\columnwidth]{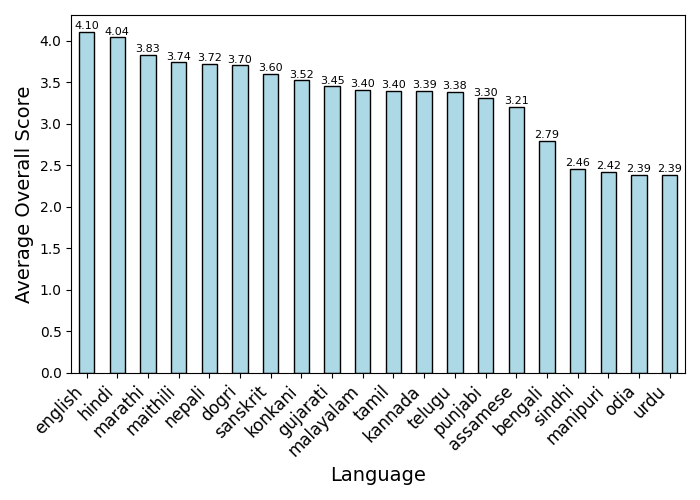}
    \caption{Language ranking based on average 'Overall' IndicQuest scores (Llama-3.1-405B-Instruct as a Judge) across languages, aggregating the scores for responses by the models. This ranking highlights the quality of multilingual LLMs for different Indic languages.}
    \label{fig:language-plot}
\end{figure}
\section{Introduction}

Language models have made tremendous progress in recent years, especially in improving performance for Indic languages \cite{gala2024airavata,team2024gemma,joshi2022l3cube}. However, the representation of these morphologically rich languages remains significantly lower compared to English and other major global languages in the current language models \cite{kakwani-etal-2020-indicnlpsuite}. This disparity exists due to a lack of large, well-structured and annotated datasets in low-resource Indic language. 

As a result of this underrepresentation, several issues such as inaccurate or inconsistent political and geographic information in Indic languages are observed frequently. For example, regional distinctions can be mistranslated, causing confusion or miscommunication. Traditional knowledge is often either contextually misrepresented or entirely omitted due to the lack of pre-training data that captures the subtleties of these languages \cite{shafayat2024multifactassessingmultilingualllms,xu2024survey}.

Addressing these disparities is important to creating inclusive language models that can represent low-resource Indic languages with the same level of sophistication as more widely spoken languages. Thus, it becomes important to quantitatively analyze the knowledge representation of these models for low-resource Indic languages, particularly when dealing with culturally and region-specific knowledge.

Current benchmarks for evaluating language models \cite{dubois2024lengthcontrolledalpacaevalsimpleway,NEURIPS2023_91f18a12} predominantly cater to English and other widely spoken languages, leaving Indic languages inadequately assessed \cite{chang2023surveyevaluationlargelanguage}. The existing benchmarks for Indic languages primarily focus on evaluating LLMs for various downstream tasks and capabilities \cite{doddapaneni2022towards} but are not suitable for assessing knowledge representation. Furthermore, there is a lack of question-answer datasets designed to evaluate these models on culturally and regionally relevant knowledge about India, which hinders their ability to evaluate the representation of Indic languages accurately.

This leaves a critical gap in assessing knowledge representation for Indic languages. To address this, we present a fact-based gold-standard Q\&A dataset for English and 19 Indic languages. This dataset is designed to evaluate how well LLMs represent Indian knowledge and will serve as a valuable resource for improving multilingual models.

Additionally, the dataset can serve two key evaluation purposes: first, for reference-based evaluations by comparing model outputs to ground truth using metrics such as ROUGE \cite{lin-2004-rouge} scores; and second, for model output assessments where a large language model (LLM) acts as the evaluator. With these approaches, the dataset facilitates a thorough evaluation of how language models handle Indic languages, offering valuable insights for future model improvements.

The key contributions of this research are as follows: 
\begin{enumerate}
    \item The development of IndicQuest, a gold-standard question-answer dataset containing 4000 questions and answers pairs, 200 each for English and the 19 Indic languages namely, Assamese, Bengali, Dogri, Gujarati, Hindi, Kannada, Konkani, Maithili, Malayalam, Marathi, Meitei (Manipuri), Nepali, Odia, Punjabi, Sanskrit, Sindhi, Tamil, Telugu, Urdu, covering five domains specific to the Indian context. This dataset is made publicly\footnote{\url{https://github.com/l3cube-pune/indic-nlp}} available.
    \item We present both reference-based evaluation and LLM-as-a-judge evaluation of various multilingual models, including GPT-4o, Llama-3.1-405B-Instruct, Llama-3.1-8B-it, Gemma-2-2B-it, and Gemma-2-9B-it, for the 19 Indic languages. Given that the judge LLM may have limitations in handling Indic facts, ground truth answers or facts are provided as references to assist the LLM in its evaluation.
    \item We demonstrate that benchmark results are stronger for English compared to the Indic languages, highlighting the disparity in knowledge representation for low-resource languages.
    \item We evaluate LLM responses against our dataset's ground truth to establish performance hierarchies across models, domains, and languages. Model ranking, based on combined evaluation metrics, is GPT-4o > Llama-3.1-405B-Instruct > Gemma-2-9B-it > Llama-3.1-8B-it > Gemma-2-2B-it. Based on the overall LLM evaluator scores,the domain ranking is Economics > Politics > History > Literature > Geography, while language ranking from highest to lowest is English, Hindi, Marathi, Maithili, Nepali, Dogri, Sanskrit, Konkani, Gujarati, Malayalam, Tamil, Kannada, Telugu, Punjabi, Assamese, Bengali, Sindhi, Manipuri, Odia, Urdu.
 (Figure \ref{fig:language-plot}).
\end{enumerate}

\section{Related Work}

TyDi QA\footnote{https://github.com/google-research-datasets/tydiqa} is a widely used question-answering benchmark that includes 11 typologically diverse languages, such as Bengali, Hindi, and Marathi, representing a variety of linguistic features. \cite{10.1162/tacl_a_00317} The dataset focuses on information-seeking questions that are naturally generated by native speakers, making it a robust benchmark for evaluating LLMs in low-resource languages. However, while TyDi QA includes several Indic languages, its primary emphasis is on typological diversity rather than region-specific contexts, which are crucial for more nuanced evaluations within specific linguistic regions like India.

XQuAD\footnote{https://github.com/google-deepmind/xquad} is a more comprehensive cross-lingual benchmark comprising 240 paragraphs and 1190 question-answer pairs from SQuAD v1.1, translated into ten languages by professional translators \cite{Artetxe_2020}.

MLQA\footnote{https://github.com/facebookresearch/MLQA} contains QA instances in seven languages: English, Arabic, German, Spanish, Hindi, Vietnamese, and Simplified Chinese. MLQA has over 12K instances in English and 5K in each other language, with each instance being parallel between four languages on average. \cite{lewis2020mlqaevaluatingcrosslingualextractive}  While MLQA includes some Indic languages, its domain and regional specificity are limited, making it less suited for a comprehensive evaluation of knowledge specific to the Indian subcontinent.

The primary application of both XQuAD and MLQA is the evaluation of question-answering capabilities of LLMs, as opposed to knowledge evaluation.

IndicQA\footnote{https://huggingface.co/datasets/ai4bharat/IndicQA} \cite{doddapaneni2022towards} is one of the few datasets explicitly targeting the evaluation of LLMs in Indic languages. It is used for evaluating question-answering models in 11 Indic languages. The context paragraphs are selected from Wikipedia articles on topics closely related to Indic culture and history. The dataset consists of 18,579 questions, of which 13,283 are answerable.  
Another recent, IndicQA benchmark \cite{singh2024indicqabenchmarkmultilingual} also focuses on evaluating closed question-answering capabilities, particularly in Indic languages. In contrast, our work addresses open-domain Q\&A without a context passage.

\section{Dataset Curation}
\subsection{Dataset Preparations}

We developed the IndicQuest dataset, a gold-standard collection of question-and-answer pairs, designed as a benchmark to evaluate the knowledge representation of Large Language Models (LLMs) in the Indian context. The dataset encompasses Q\&As in English and 19 major Indic languages: Assamese, Bengali, Dogri, Gujarati, Hindi, Kannada, Konkani, Maithili, Malayalam, Marathi, Meitei (Manipuri), Nepali, Odia, Punjabi, Sanskrit, Sindhi, Tamil, Telugu, Urdu.

For dataset curation, we formulated factual question-and-answer pairs in English, sourced from reputable platforms like Wikipedia and well-known educational websites.
The questions were structured across five key domains Literature, History, Geography, Politics, and Economics based on resource availability, topic importance, and cultural relevance to the Indian context. Each domain consists of 40 questions, totaling 200 per language. History, Geography, and Politics questions cover specific sub-regions of India, ensuring representation of the northern, eastern, western, and southern belts. Economics questions address national-level topics, while Literature questions are split between Western and Indian literary works familiar to the Indian audience.

A thorough manual verification process was conducted to ensure the accuracy of the English dataset by cross-referencing answers with reliable sources to eliminate ambiguity. The verified question-answer pairs were then translated into 19 Indic languages using Google Translate, maintaining the linguistic accuracy, of the languages.

\begin{figure}[H]
    \centering
    \includegraphics[width=\columnwidth]{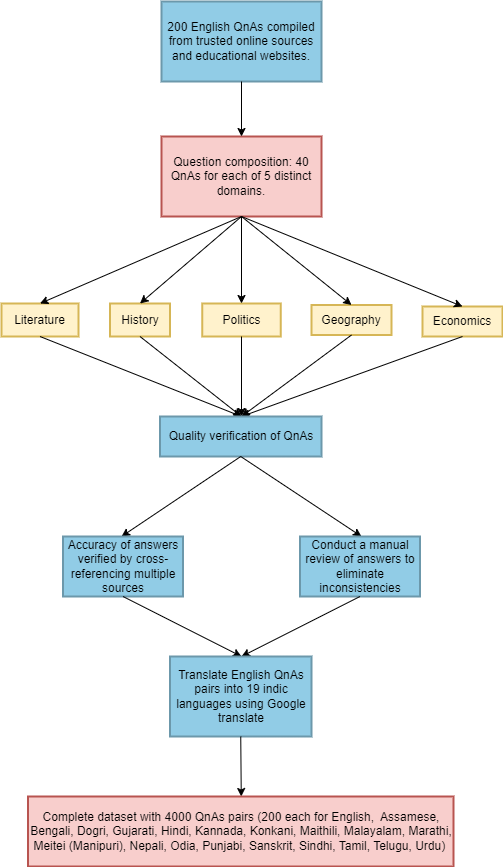}
    \caption{Dataset Curation Process}
    \label{fig:dataset-curation-flowchart}
\end{figure}

\begin{figure}[H]
    \centering    \includegraphics[width=\columnwidth]{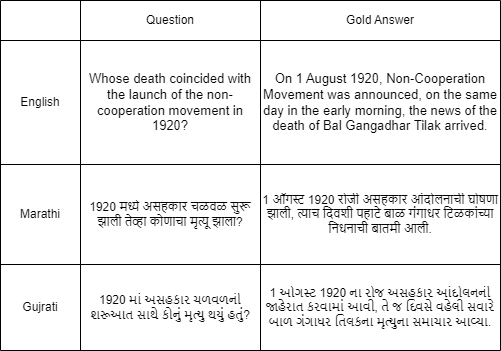}
    \caption{Dataset Overview}
    \label{fig:dataset-overview}
\end{figure}

\subsection{Data Statistics}
\begin{itemize}
    \item \textbf{Total Q\&As:} 4000 (200 questions per language)
\item \textbf{Domains:} 5 (Literature, History, Geography, Politics, Economics)
\item \textbf{Language Distribution:} Equal distribution across 20 languages (English + 19 Indic languages)
\item \textbf{Domain Distribution:} 40 questions per domain per language.
Sub-regional coverage: Balanced representation of northern, eastern, western, and southern regions in History, Geography, and Politics.
\end{itemize}
% \begin{figure}[H]
%     \centering
%     \includegraphics[width=\columnwidth]{flowchart.png}
%     \caption{Dataset Curation Process}
%     \label{fig:dataset-curation-flowchart}
% \end{figure}
% \begin{figure}
%     \centering    \includegraphics[width=\columnwidth]{table.drawio.png}
%     \caption{Dataset Overview}
%     \label{fig:dataset-overview}
% \end{figure}

\section{Evaluation Methodology}

We conducted an evaluation of the knowledge representation capabilities of various Large Language Models (LLMs) using our IndicQuest dataset as a benchmark. The evaluation covered a diverse set of LLMs, including both proprietary and open-source models, across various sizes. The models evaluated were: Gemma-2-2b-it\footnote{https://huggingface.co/google/gemma-2-2b-it}, Gemma-2-9b-it\footnote{https://huggingface.co/google/gemma-2-9b-it}, Llama-3.1-8b-it\footnote{https://huggingface.co/meta-llama/Meta-Llama-3.1-8B}, Llama-3.1-405b-Instruct\footnote{https://huggingface.co/meta-llama/Llama-3.1-405B-Instruct} and GPT-4o. Model responses were generated for  English and the 19 Indic language gold standard questions, and systematically compared to the corresponding gold standard answers (ground truth) in our dataset. Due to limited resources, GPT-4o responses were obtained only for English, Marathi, and Hindi. The evaluation utilized three distinct performance metrics to assess the degree of alignment between the model-generated responses and the ground truth answers. The results of the evaluation are shown in Table~\ref{tab:model-evaluation}. 
% \\
\subsection{Evaluation Metrics}
To assess the quality of the responses, we employed the following metrics: 

\begin{enumerate}
    \item \textbf{Automated Evaluation with Llama-3.1-405B-Instruct (LLM as a Judge):} We utilized the Llama-3.1-405B-Instruct\footnote{https://huggingface.co/meta-llama/Meta-Llama-3.1-405B-Instruct} model to automatically evaluate the responses generated by the aforementioned models. The evaluation was guided by a structured prompt provided to Llama as shown in Listing 1, specifying five key criteria: Factual Accuracy, Relevance, Clarity, Language Consistency, and Conciseness. Each criterion was explicitly outlined in the prompt to ensure a consistent evaluation approach. The model assigned scores to the responses on a scale of 1.0 to 5.0 for each criterion. Additionally, the prompt instructed Llama-3.1-405B-Instruct to calculate an Overall score (also on a scale of 1.0 to 5.0) considering these five key criteria scores, with Factual Accuracy being given a higher weightage. This Overall score, is reported in Table~\ref{tab:model-evaluation}.

It is important to note that the automated evaluation using Llama-3.1-405B-Instruct was not performed for answers generated by Llama-3.1-405B-Instruct itself to avoid potential bias.

\lstset{
  language=Python,
  basicstyle=\tiny\ttfamily, % Adjust font size here
  keywordstyle=\color{blue},
  commentstyle=\color{black},
  stringstyle=\color{red},
  numbers=none,
  frame=single,
  breaklines=true,
  rulecolor=\color{black},
  linewidth=0.425\textwidth, % Adjust the width of the code block
  % xleftmargin=0.05\columnwidth, % Margin on the left side
  % xrightmargin=0.05\columnwidth % Margin on the right side
}

\begin{lstlisting} [caption={Evaluation prompt given to Llama-3.1-405b}, label={lst:prompt-code}]
prompt = f"""
Evaluate the quality of the model's responses to questions from a benchmark dataset on a scale of 1-5 (score can be a decimal fraction format number) across the following parameters:

Factual Accuracy: Given an input question, ground truth facts relevant to the question, and the model/bot's answer, evaluate how well the information in the model's answer aligns with the provided ground truth facts. Assign a score on a scale of 1 to 5 based on the following criteria: a score of 5 indicates complete alignment with all ground truth facts; a score of 3 represents partial alignment where approximately half of the facts are correct; and a score of 1 denotes complete misalignment with the ground truth facts. Scores between these benchmarks can reflect varying degrees of alignment or discrepancies.

Relevance: Assess how well the model's answer directly addresses the question. A score of 5 indicates a highly relevant answer, while a score of 1 indicates an irrelevant or off-topic response.

Clarity: Evaluate the clarity and coherence of the model's answer. A score of 5 means the answer is well-structured and easy to understand, while a score of 1 means it is confusing or poorly constructed.

Language Consistency: Ensure that the language of the response matches the language of the question unless otherwise specified. Penalize cases where there is a mismatch between the input language specified in the question and the response language.

Conciseness: Rate how concise the answer is while still providing necessary information. A score of 5 indicates the answer is succinct and to the point, while a score of 1 indicates excessive verbosity or unnecessary information.

Input Details:
Question: {question}
Ground Truth Facts: {ground_truth}
Model/Bot Answer: {model_answer}
After evaluating each parameter, provide an overall rating on a scale of 1-5 considering all the parameters. The parameter factual accuracy should have more weightage in the overall score.

Output Format:
Return the evaluation scores in the following JSON format(Return only the JSON and nothing else):
  {{
    "Factual Accuracy": score,
    "Relevance": score,
    "Clarity": score,
    "Language Consistency": score,
    "Conciseness": score,
    "Overall": average_score
  }}
"""
\end{lstlisting} 

\item \textbf{F1 Score:} This metric provided a combined measure of precision and recall to further assess the quality of the model outputs.

\item \textbf{ROUGE Score:} We calculated the ROUGE (Recall-Oriented Understudy for Gisting Evaluation) score \cite{lin-2004-rouge} to measure the overlap between the model-generated responses and the ground truth answers. 

\end{enumerate}

% \begin{figure}
%     \centering
%     \includegraphics[width=\columnwidth]{model_performance (1).png}
%     \caption{Average 'Overall' scores across Models}
%     \label{fig:model-plot}
% \end{figure}

\begin{table*}[t]
\centering
\small
\setlength{\tabcolsep}{3pt}
\renewcommand{\arraystretch}{1.5}
\resizebox{\textwidth}{!}{%
\begin{tabular}{|l|c|*{20}{c|}}
\hline
\multirow{3}{*}{\normalsize\textbf{Model}} & \multirow{3}{*}{\normalsize\textbf{Metric}} & \multicolumn{20}{c|}{\normalsize\textbf{Language Scores}} \\
\cline{3-22}
 &  & En & Hi & Mr & Mi & Ne & Do & Sa & Ko & Ml & Ta & Ka & Te & Pu & As & Be & Si & Od & Ur & Gu & Mn \\
\hline
\multirow{3}{*}{\makecell{Gemma-2-2B-it}} 
 & Overall Score & 3.81 & 3.55 & 3.28 & 3.34 & 3.32 & 3.29 & 3.24 & 3.10 & 2.82 & 3.00 & 2.80 & 2.70 & 2.55 & 2.58 & 2.38 & 1.95 & 1.66 & 1.66 & 2.79 & 1.83 \\
 & F1 & 0.61 & 0.47 & 0.33 & 0.30 & 0.29 & 0.23 & 0.16 & 0.23 & 0.24 & 0.26 & 0.27 & 0.26 & 0.29 & 0.23 & 0.02 & 0.11 & 0.15 & 0.34 & 0.31 & 0.01 \\
 & ROUGE-L & 0.18 & 0.08 & 0.06 & 0.03 & 0.05 & 0.06 & 0.01 & 0.03 & 0.06 & 0.05 & 0.06 & 0.08 & 0.06 & 0.01 & 0.06 & 0.04 & 0.01 & 0.04 & 0.07 & 0.00 \\
\hline
\multirow{3}{*}{\makecell{Gemma-2-9B-it}}
 & Overall Score & 4.17 & 4.11 & 3.98 & 4.05 & 4.14 & 4.03 & 4.06 & 3.83 & 3.95 & 3.90 & 3.94 & 3.92 & 3.79 & 3.78 & 3.28 & 2.86 & 2.70 & 2.70 & 3.93 & 2.26 \\
 & F1 & 0.63 & 0.57 & 0.44 & 0.44 & 0.41 & 0.35 & 0.23 & 0.29 & 0.34 & 0.37 & 0.34 & 0.38 & 0.47 & 0.35 & 0.03 & 0.27 & 0.26 & 0.51 & 0.43 & 0.18 \\
 & ROUGE-L & 0.21 & 0.10 & 0.03 & 0.07 & 0.04 & 0.08 & 0.01 & 0.02 & 0.09 & 0.10 & 0.10 & 0.10 & 0.10 & 0.01 & 0.06 & 0.09 & 0.06 & 0.11 & 0.10 & 0.00 \\
\hline
\multirow{3}{*}{\makecell{Llama-3.1-8B-it}}
 & Overall Score & 3.98 & 4.01 & 3.78 & 3.83 & 3.71 & 3.78 & 3.49 & 3.63 & 3.44 & 3.29 & 3.44 & 3.52 & 3.58 & 3.25 & 2.72 & 2.58 & 2.79 & 2.79 & 3.62 & 3.17 \\
 & F1 & 0.60 & 0.49 & 0.34 & 0.37 & 0.31 & 0.31 & 0.19 & 0.26 & 0.24 & 0.23 & 0.24 & 0.27 & 0.44 & 0.25 & 0.02 & 0.29 & 0.19 & 0.42 & 0.33 & 0.24 \\
 & ROUGE-L & 0.19 & 0.11 & 0.07 & 0.06 & 0.04 & 0.08 & 0.01 & 0.04 & 0.04 & 0.08 & 0.05 & 0.08 & 0.10 & 0.00 & 0.04 & 0.08 & 0.07 & 0.12 & 0.07 & 0.00 \\
\hline
\multirow{3}{*}{\makecell{Llama-3.1-405B-it}}
 & Overall Score & - & - & - & - & - & - & - & - & - & - & - & - & - & - & - & - & - & - & - & - \\
 & F1 & 0.67 & 0.55 & 0.39 & 0.42 & 0.38 & 0.39 & 0.27 & 0.32 & 0.30 & 0.33 & 0.33 & 0.33 & 0.55 & 0.35 & 0.03 & 0.38 & 0.29 & 0.51 & 0.43 & 0.26 \\
 & ROUGE-L & 0.23 & 0.13 & 0.07 & 0.09 & 0.05 & 0.11 & 0.00 & 0.07 & 0.08 & 0.09 & 0.06 & 0.08 & 0.13 & 0.00 & 0.01 & 0.15 & 0.10 & 0.13 & 0.06 & 0.00 \\
\hline
\multirow{3}{*}{GPT-4o}
 & Overall Score & 4.45 & 4.49 & 4.27 & - & - & - & - & - & - & - & - & - & - & - & - & - & - & - & - & - \\
 & F1 & 0.70 & 0.73 & 0.53 & - & - & - & - & - & - & - & - & - & - & - & - & - & - & - & - & - \\
 & ROUGE-L & 0.23 & 0.13 & 0.08 & - & - & - & - & - & - & - & - & - & - & - & - & - & - & - & - & - \\
\hline
\end{tabular}
}
\caption{Evaluation scores for all models across 20 languages. Language abbreviations: En=English, Hi=Hindi, Mr=Marathi, Mi=Maithili, Ne=Nepali, Do=Dogri, Sa=Sanskrit, Ko=Konkani, Ml=Malayalam, Ta=Tamil, Ka=Kannada, Te=Telugu, Pu=Punjabi, As=Assamese, Be=Bengali, Si=Sindhi, Od=Odia, Ur=Urdu, Gu=Gujarati, Mn=Manipuri. Dashed entries (-) indicate scores not obtained for those particular languages, as GPT-4o responses were generated only for En, Hi, and Mr, while Llama-3.1-405b responses were evaluated only using F1 and ROUGE metrics.}
\label{tab:model-evaluation}
\end{table*}

\begin{figure}[H]
    \centering
    \includegraphics[width=\columnwidth]{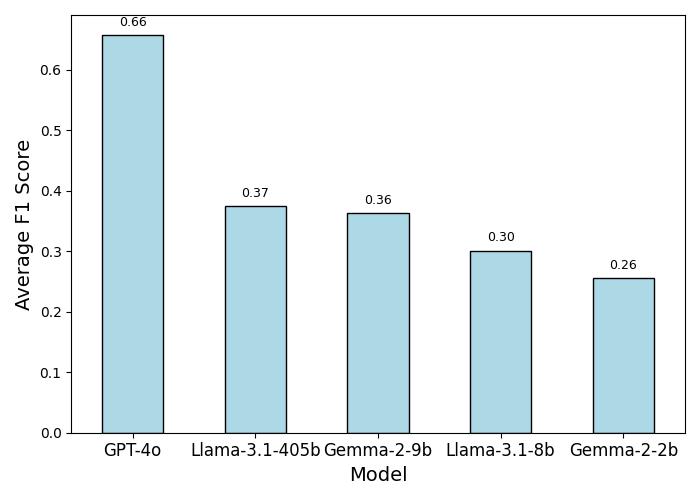}
    \caption{Average F1 Scores across Models obtained by aggregating the scores for all responses to Questions in IndicQuest given by these models. This ranking highlights model performance for Indic languages.  }
    \label{fig:model-plot}
\end{figure}
\begin{figure}[H]
    \centering
    \includegraphics[width=\columnwidth]{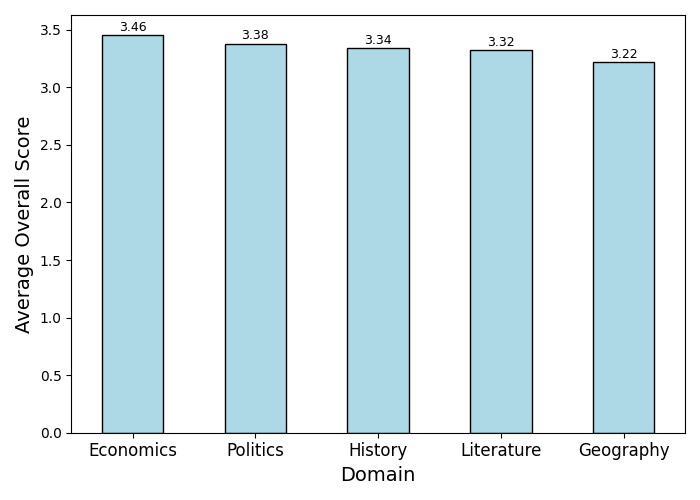}
    \caption{Average 'Overall' scores across Domains obtained by aggregating the scores for responses of all languages and models for the domain. This indicates model performance across various domains.}
    \label{fig:domain-performance}
\end{figure}
\section{Results and Observations}
The following observations were made from the obtained metric scores after our evaluation:

\begin{enumerate}
\item \textbf{GPT-4o Outperforms across the languages, with Larger Models Leading in Performance:} There is a clear hierarchy in model performance across most domains and languages, with GPT-4o consistently outperforming the other models, followed by Llama-3.1-405B-Instruct, Gemma-2-9B-it, Llama-3.1- 8B-it, and Gemma-2B-it (Figure~\ref{fig:model-plot}). As shown in Table~\ref{tab:model-evaluation}, models with larger parameter counts consistently achieve better results, reinforcing the correlation between model size and performance. This hierarchy was determined through analysis of Overall Llama score, F1, and ROUGE scores.
\item \textbf{Stronger English Performance Validates the Need for Greater Representation of Indic Languages in Multilingual LLMs:} All models demonstrate significantly stronger performance in English compared to the Indic languages, as evident in Figure~\ref{fig:language-plot}, which shows a clear performance hierarchy with English at the top, followed by the indic languages that show relatively lower performance. We can see a hierarchy in the language performance in Figure~\ref{fig:language-plot} where the average scores were obtained by considering the scores for the responses evaluated using Llama-3.1-405B-Instruct as a judge. This disparity is consistent across all models and highlights a gap in multilingual proficiency, especially for low-resource Indic languages. These findings reinforce the initial motivation of this study: to increase the representation of Indic languages in multilingual LLMs. 
% The evaluation metrics Llama Score, ROUGE-L, and F1 further support these findings, reflecting consistent trends across languages and domains.
% \begin{figure}
%     \centering
%     \includegraphics[width=\columnwidth]{language_performance_gemma9b (1).png}
%     \caption{Average 'Overall' scores across languages, calculated specifically for Gemma-2-9B-it.}
%     \label{fig:language-plot}
% \end{figure}
\item \textbf{Domain Performance Disparities Reflect Gaps in Region-Specific Knowledge:} The models exhibit performance variation across different domains, suggesting that certain areas of region-specific knowledge are unevenly represented in the LLMs. Based on the Overall Score, a clear hierarchy emerges with Economics performing best, followed by Politics, History, Literature, and Geography being the weakest domain across all languages. These disparities imply that some domains may lack sufficient pre-training data or require domain-specific fine-tuning to improve results. This highlights the need for more focused training on culturally and regionally relevant content in multilingual models.
% \begin{figure}
%     \centering
%     \includegraphics[width=\columnwidth]{domain_performance (1).png}
%     \caption{Average 'Overall' scores across Domains}
%     \label{fig:domain-performance}
% \end{figure}
\item \textbf{Need for Improvement in Multilingual Capabilities Despite Superior Performance of GPT-4o:} While GPT-4o consistently outperforms smaller models across English, Marathi and Hindi datasets, its performance in Marathi still lags behind its English counterpart. This discrepancy highlights the limitations of even the most advanced models when it comes to low-resource languages like Marathi and other Indic languages.
\end{enumerate}

\section{Conclusion and Future Work}
In this work, we introduce IndicQuest, a resource designed to evaluate Large Language Models (LLMs) for their ability to represent knowledge in Indic languages. The dataset comprises 4,000 gold-standard question-answer pairs, with 200 pairs each for English and 19 Indic languages. We conducted an evaluation for English and Marathi on various multilingual models, using Llama-3.1- 405b-it as the evaluator, alongside standard metrics such as ROUGE and F1 score. Our evaluation involved five models: Gemma-2-2b-it, Gemma-2-9b-it, Llama-3.1-8b-it, Llama-3.1-405b-Instruct and GPT-4o.

From the evaluation scores, we observed a disparity in the performance of LLMs between English and the Indic languages. Despite advancements, English—a well resourced language continues to outperform Indic languages. This underscores the need for further improvements in LLMs to enhance their inclusion of Indic languages, as well as the importance of developing Indic knowledge-based benchmark datasets to identify areas where these models fall short in Indic-specific contexts.

	The evaluation process in this study was fully automated. In the future, we plan to conduct human evaluations on the English, Marathi and Hindi subsets, involving subject matter experts, to compare their assessments with the automated Llama Evaluation results. Additionally, we aim to perform a deep quantitative analysis of the results to identify specific linguistic and domain-related challenges faced by the models.
	
 We hope this dataset will serve as a valuable benchmark for advancing research in multilingual LLMs, particularly in evaluating their performance and using it as a standard for assessment.

\section{Acknowledgements}
This work was carried out under the mentorship program of L3Cube Labs, Pune. We would like to express our sincere gratitude to our mentor, for his continuous support and guidance.

\bibliography{main.bib}

\end{document}